\documentclass{SCIS2025}
\usepackage{multirow}
\usepackage{tikz}

\usepackage{mdframed}
\usepackage{xcolor}

\begin{document}

\ArticleType{RESEARCH PAPER}
\Year{2024}
\Month{}
\Vol{}
\No{}
\DOI{}
\ArtNo{}
\ReceiveDate{}
\ReviseDate{}
\AcceptDate{}
\OnlineDate{}
\title{Fine-grained text-driven dual-human motion generation via dynamic hierarchical interaction}{Fine-grained text-driven dual-human motion generation via dynamic hierarchical interaction}
\author[1]{Mu Li}{}
\author[1]{Yin Wang}{}
\author[1,]{Zhiying Leng}{{zhiyingleng@buaa.edu.cn}}
\author[1]{Jiapeng Liu}{}
\author[2]{Frederick W. B. Li}{}
\author[1,3]{Xiaohui Liang}{}

\AuthorMark{Mu Li, Yin Wang}

\AuthorCitation{Mu Li, Yin Wang, et al}

\address[1]{State Key Laboratory of Virtual Reality Technology and Systems, Beihang University, Beijing, China}
\address[2]{Department of Computer Science, University of Durham, UK}
\address[3]{Zhongguancun Laboratory, Beijing, China}

\abstract{Human interaction is inherently dynamic and hierarchical, where the dynamic refers to the motion changes with distance, and the hierarchy is from individual to inter-individual and ultimately to overall motion. Exploiting these properties is vital for dual-human motion generation, while existing methods almost model human interaction temporally invariantly, ignoring distance and hierarchy. To address it, we propose a fine-grained dual-human motion generation method, namely FineDual, a tri-stage method to model the dynamic hierarchical interaction from individual to inter-individual. The first stage, Self-Learning Stage, divides the dual-human overall text into individual texts through a Large Language Model, aligning text features and motion features at the individual level. The second stage, Adaptive Adjustment Stage, predicts interaction distance by an interaction distance predictor, modeling human interactions dynamically at the inter-individual level by an interaction-aware graph network. The last stage, Teacher-Guided Refinement Stage, utilizes overall text features as guidance to refine motion features at the overall level, generating fine-grained and high-quality dual-human motion. Extensive quantitative and qualitative evaluations on dual-human motion datasets demonstrate that our proposed FineDual outperforms existing approaches, effectively modeling dynamic hierarchical human interaction.}


\maketitle

\section{Introduction}

Human motion generation synthesizes realistic motions, widely applied in augmented/virtual reality, animation, gaming, and film visualization. While existing works have explored modalities such as music~\cite{kao2020temporally,li2021ai,ren2020self,starke2022deepphase,tseng2023edge}, actions~\cite{guo2020action2motion,petrovich2021action,cervantes2022implicit}, and trajectories~\cite{karunratanakul2023guided,shafir2023human,wan2023tlcontrol} for human motion generation, text-driven approaches~\cite{guo2022generating,petrovich2022temos,tevet2023human,chen2023executing,zhang2022motiondiffuse,wang2023fg,bhattacharya2021text2gestures,jin2024act,zhang2024finemogen,zhang2024large,wang2025fg,wang2025most} have received interest due to their semantic expressivity and usability. However, most of the existing methods are dedicated to single-human motion generation, neglecting the dual-human motion that often occurs. Dual-human motion generation poses substantial challenges compared to single-human motion due to the complex dual-human interaction that emerges, such as the fight between boxers, the physical contact between dancers, and so on.

\begin{figure*}[t]
    \centering
    \includegraphics[width=\linewidth]{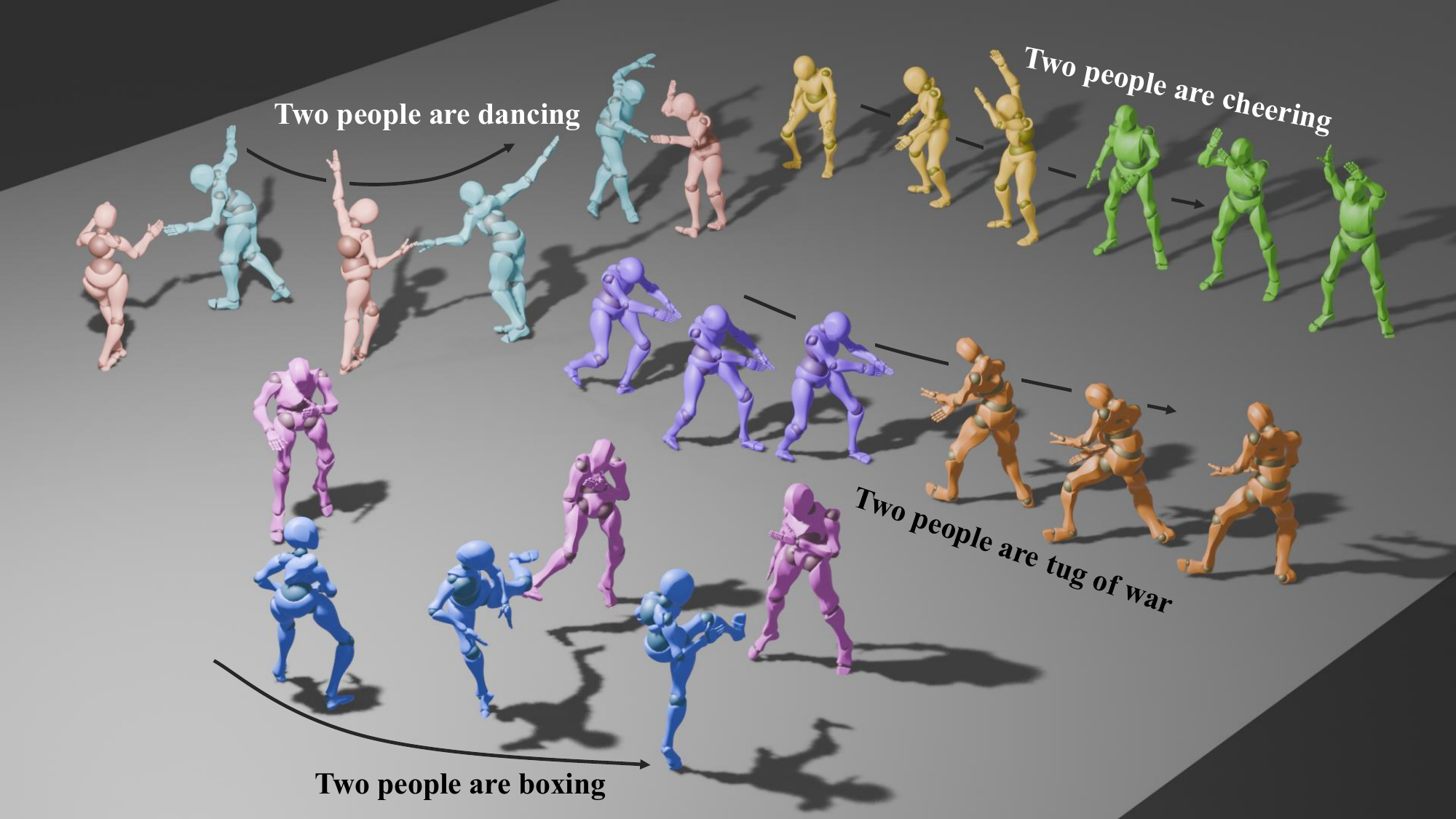}
    \caption{We propose FineDual to generate high-quality dual-human motions with fine-grained interactions. Here, we present four text prompts along with their corresponding motions. Each color represents an individual. The arrow represents the time axes.}
    \label{teaser}
\end{figure*}

Recently, a few works have appeared to explore dual-human motion generation preliminarily. Shafir \textit{et al.}~\cite{shafir2023human} employed the transformer layer as a communication module to model motion interaction. InterGen~\cite{liang2024intergen} simply used cross-attention on the entire motion to establish interaction between two individuals. Additionally, DLP~\cite{cai2023digital} introduced motion-matching technology in its cross-attention module to boost interaction quality. However, these methods model human interaction temporally invariantly, leading to generating coarse dual-human motions, such as unreasonable human interaction.

Insights from the concept of proxemics inform our research. 
In The Hidden Dimension~\cite{hall1966hidden}, Edward T. Hall introduced proxemics, emphasizing the strong correlation between physical distance and interpersonal behavior. Hall notes that: ``Social distance between people is reliably correlated with physical distance," highlighting the close connection between social interactions and spatial proximity.

Based on this, we observe that the inherent properties of real-world human interactions—dynamic and hierarchy—are beneficial for generating fine-grained dual-human motions.
Firstly, dynamic refers to that ``\emph{the interactive motions between two individuals are inversely related to their distance}". Specifically, as the physical distance between individuals decreases, people generally exhibit increased attentiveness to their partner’s motions. For example, in a boxing sports scenario, when two people are far apart, they focus on performing their own motions, whereas at a close distance, they respond based on the other's motions.
Secondly, hierarchy means that the human interaction is hierarchical, ``\emph{root is individual motion, the middle is inter-individual interaction, the upper is overall motion}". 
Based on these properties, we propose a text-driven fine-grained dual-human motion generation method that models the dynamic hierarchical human interaction. 

In detail, our proposed FineDual, a hierarchical tri-stage approach, models dynamic human interactions from individual to inter-individual, and ultimately to overall motions. Firstly, the \textbf{self-learning stage} initiates the hierarchy from the individual level. The stage utilizes Large Language Models (LLMs) to break down the overall prompt into individual prompts, thereby aligning the text and motion features of each individual independently. Following this, the \textbf{adaptive adjustment stage} as the middle inter-individual level of the hierarchy, models inter-individual dynamic interactions through the dual-human graph structure. In particular, we propose an interaction distance predictor to dynamically predict the interactive distance as the edge weight of the graph, in which interactive distance indicates the Euclidean distance between individuals. Under the impact of the interaction distance, our proposed interaction-aware graph reasoning module injects dynamic human interaction into motion features. Finally, the \textbf{teacher-guided refinement stage} as the upper overall level, utilizes overall text prompts as guidance to refine motion features, allowing FineDual to model fine-grained interaction by multi-modal conditions. In a word, our proposed FineDual models human interaction in a dynamic hierarchical way.

To validate the effectiveness of our method, we conducted experiments on the existing public dual-human dataset, InterHuman \cite{liang2024intergen} and Inter-X \cite{xu2024inter}, and proposed new metrics to evaluate dual-human generation. The results demonstrate that our method has achieved superior performances compared to existing methods. As Figure \ref{teaser} depicts, FineDual addresses producing fine-grained dual-human motions guided by textual prompts through interaction modeling. Our contributions are as follows:

\begin{itemize}

\item We propose a tri-stage dynamic hierarchical interaction framework for dual-human motion generation, which precisely models the fine-grained interactions between individuals.
    
\item We propose to explicitly utilize distance information to model human interaction, capturing how distance impacts subtle dual-human interaction dynamics.

\item We present several new evaluation metrics for evaluating human-human interactions. Experimental results validate that FineDual achieves state-of-the-art performance on existing datasets.

\end{itemize}

\section{Related work}

\subsection{Text-driven single-human motion generation}
Three primary methodologies have emerged to tackle the challenge of text-driven single-human motion generation. (i) Latent space alignment, exemplified by JL2P \cite{ahuja2019language2pose} and TEMOS \cite{petrovich2022temos}, aims to learn a unified latent space between textual and motion embeddings. (ii) Conditional autoregressive models produce motion tokens in sequence, drawing on prior tokens and texts. Noteworthy contributions in this domain include TM2T \cite{guo2022tm2t}, which employs a vector quantized VAE, and T2M-GPT \cite{zhang2023generating}, which refines motion tokens through exponential moving averages and code resetting. PoseGPT \cite{lucas2022posegpt} proposed an auto-regressive transformer-based approach to quantize human motion into latent sequences, generating realistic and diverse 3D human motions from given the human action, a duration, and an arbitrarily long past observation. Additionally, MoMask \cite{guo2023momask} and MMM \cite{pinyoanuntapong2023mmm} employ a masked motion model for more natural motion generation. (iii) Conditional diffusion models, such as MotionDiffusion \cite{zhang2022motiondiffuse}, Fg-T2M \cite{wang2023fg}, MLD \cite{chen2023executing}, and MDM \cite{tevet2023human} have shown superior performance by leveraging conditional diffusion framework to learn probabilistic text-motion mappings. While these advancements have propelled human motion generation forward, they predominantly center on individual motion generation, lacking the ability to generate interactive motions involving two or more individuals.

\subsection{Text-driven dual-human motion generation} 

In a notable effort, ComMDM \cite{shafir2023human} harnessed two pre-trained MDM \cite{tevet2023human} models as generative priors and incorporated a communication module to orchestrate motions for pairs of individuals. Tanaka \textit{et al.}\cite{tanaka2023role} proposed learning the pair motions by distinguishing the dual-human motions into actor and receiver, interaction modeling of which is also achieved by a shared cross-attention module. DiffuGesture \cite{zhao2023diffugesture} integrates a lightweight transformer encoder to harmonize temporal dynamics between human gestures and multi-modal conditions, generating dual-human gesture motions. in2IN \cite{Ruiz-Ponce_2024_CVPR} employs both interaction-level and individual-level textual descriptions to jointly facilitate motion generation. Meanwhile, in2IN introduces a diffusion conditioning technique that independently weights the importance of each condition. 
However, this approach fails to account for mutual influence between individuals and lacks explicit modeling of human interactions, resulting in only sub-optimal dual-human motions. 
InterGen \cite{liang2024intergen} introduced dual collaborative denoisers that share weights and establish links through mutual attention mechanisms. DLP \cite{cai2023digital} innovatively melds reflective processes grounded in psychological principles through SocioMind and steers dual-human motion generation by MoMat-MoGen. 
However, these methods typically use temporally invariant interaction modeling approaches, ignoring interaction distance between individuals. Therefore, their generation quality is sub-optimal and does not effectively convey human-to-human interactions. In this paper, we propose FineDual, a fine-grained dynamic hierarchical interaction method. FineDual explicitly utilizes distance information to model human interaction to refine the interaction process through three stages in dual-human scenarios. 

\subsection{Distance information in human motion tasks}
Distance information has been preliminarily attempted in human motion forecasting and trajectory prediction tasks. In human motion prediction, Tang \textit{et al.} \cite{tang2018long} proposed a loss function to supervise the skeleton distance between two frames, in order to generate a smooth motion sequence. Zhu \textit{et al.} \cite{zhu2024development} proposed the discrete control barrier function (DCBF) to effectively predict human motion. DCBF is redefined as the distance between each link of the robot and each part of the human as the line distance. In trajectory prediction, Habibi \textit{et al.} \cite{habibi2018context} incorporated semantic features of the environment, such as the distance to the curbside and the status of pedestrian traffic lights, into the Gaussian Process (GP) formulation to more accurately predict trajectories. Batz \textit{et al.} \cite{batz2009recognition} predicted the distance among traffic participants to identify dangerous situations in the field of vehicle trajectory prediction. If the distance is below the predefined threshold, a dangerous situation is identified.
However, how to use distance information in dual-human motion generation is still unexplored. In this work, we propose to utilize the distance between humans to explicitly model human interaction.

\section{Methodology}

\begin{figure*}[t]
    \centering
    \includegraphics[width=\linewidth]{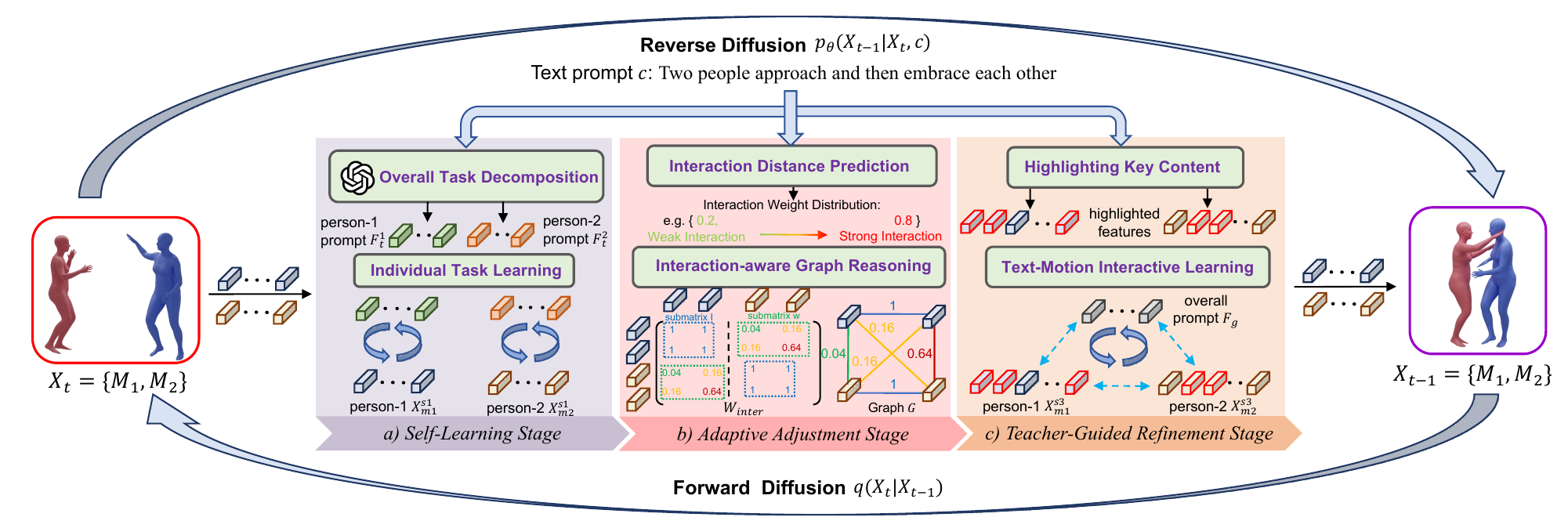}
    \caption{\textbf{Overview of Our FineDual.} The dynamic hierarchical interaction models human interactions at progressively deeper levels. Given the text prompt $c$, our method refines motion from noisy data through T steps, with each stage addressing a distinct level of interaction understanding. Taking step $t$ as an example: (1) The \textit{Self-Learning Stage} captures individual motion features for each person separately based on their own text prompts, producing $X_{m1}^{s1}$ and $X_{m2}^{s1}$. (2) In the \textit{Adaptive Adjustment Stage}, the model builds an interaction-aware graph to assess the relationship between individuals, predicting interaction strength and assigning weights to adjust motion features, resulting in interaction-refined outputs $X_{m1}^{s2}$ and $X_{m2}^{s2}$. (3) The \textit{Teacher-Guided Refinement Stage} synthesizes these individual and interaction-level features with the overall prompt context, generating keyframe-adjusted, final motion outputs $X_{m1}^{s3}$ and $X_{m2}^{s3}$ that represent cohesive and contextually appropriate interactions.}
    \label{pipeline}
\end{figure*}

\subsection{Motion generation via diffusion models}

Given the text prompts as the condition, we propose a conditional diffusion model based method, FineDual, generating fine-grained dual-human motion sequences from random noise. As shown in Figure \ref{pipeline}, our FineDual consists of three stages, the self-leanring stage described in Section~\ref{als}, the adaptive adjustment stage shown in Section~\ref{aas}, and the teacher-guided refinement stage introduced in Section~\ref{tgrs}. Formally, given the random noisy dual-human motion close to the random Gaussian noise, $X_{T}=\{M_{i} |M_i \in \mathbb{R}^{S \times D} \}$, our FineDual takes the text prompt $c$ as conditions to generate the clean dual-human motion $X_{0}$ backward by T steps. Here, $i$ indicates person $i$, $S$ refers to the motion sequence length, and $D$ is the dimension of each person's motion representation.

To supervise this motion generation process, we minimize the L2 loss between predicted and ground truth motions, denoted as 
\begin{equation}
\mathcal{L}_{1}=\mathbb{E} [\parallel \mathbf{x}_0 - \epsilon_\theta(\mathbf{x}_t,t,c) \parallel_2^2]. 
\end{equation}
where $\epsilon_\theta(\mathbf{x}_t,t,c)$ denotes the model prediction. Additionally, we employ classifier-free diffusion guidance \cite{ho2022classifier} to scale conditional and unconditional distributions as:
\begin{equation} 
\epsilon = s\epsilon_\theta(x_t,t,c)+(1-s)\epsilon_\theta(x_t,t,\varnothing).
\end{equation}
where $t$ denotes the timestep, $c$ and $\varnothing$ represent using text condition or not, respectively. Guidance scale $s$ controls the strength of the text condition.

\subsection{Self-learning stage} \label{als}

In human interaction, the individual's own motion quality affects the overall interaction process. Drawing inspiration from this observation, individual self-learning in dual-human motion is a vital factor in interaction between individuals. 
However, current methods tend to directly derive limiting individual features from the given overall text prompts, which are often confined to generating coarse-grained interactive motions, such as ``one person kicked another with his left leg." Such oversimplified patterns present two primary challenges. Firstly, overall prompts fail to align with each individual, resulting in reduced quality and completeness of the individual's specific motions. Secondly, directly modeling dual-human motions accurately with overall and coarse-grained prompts is challenging.

Hence, we propose a self-learning stage dedicated to learning individual features separately at the individual level. Only after individuals have mastered their foundational tasks should we furthermore proceed to more complex tasks that involve interactions between individuals. As shown in Figure \ref{pipeline} a), the self-learning stage encompasses two steps, overall task decomposition for parsing the overall text prompts into detailed descriptions for each individual, and individual task learning based on individual text prompts.

\textbf{Overall task decomposition. }
State-of-the-art large language models (LLMs), such as OpenAI's GPTs \cite{achiam2023gpt}, have revolutionized the NLP landscape with their robust modeling capabilities. We leverage the advanced prior knowledge of GPT-3.5 to deconstruct the overall text prompt $T_g$ into tailored, individual prompts $T_t^1$ and $T_t^2$. This approach enables each individual to learn from a personalized text prompt, bridging the divide between precise motion data and model-generated motions. Specifically, our designed prompt is as follows:

\begin{mdframed}[linewidth=0.5pt,roundcorner=10pt,backgroundcolor=gray!10]
\textsl{\textbf{Prompt}: We have a series of text descriptions that depict motions involving two people. I would like to divide the sentence into the respective motions of these two individuals. The goal is to analyze and produce a list, which contains two key-value pairs: person1 and person2. 
There are generally three types of scenarios:}
\begin{enumerate}
\item \textsl{The sentence only describes the overall motions of two people without specifying each person's individual motion. In this case, the values for person1 and person2 are directly the original text. }

\textsl{\textbf{Example}: these two return to their original position.  }

\textsl{ \textbf{Answer}:  \{person1: he returns to his original position, person2: he returns to his original position.\} }

\item \textsl{The sentence describes the motions of two people and uses terms like ``one, the other" or ``the first person, the second" to distinguish between them. The values for person1 and person2 correspond to these individuals respectively. }

\textsl{\textbf{Example}: one person is crossing the legs, the other person takes a picture.}

\textsl{\textbf{Answer}: \{person1: one person is crossing the legs, person2: the other person takes a picture.\}}

\item \textsl{The sentence describes the motions of two people but does not explicitly describe the second person's action. The value for person1 is the original sentence, while the value for person2 needs to be generated based on the context to provide a reasonable motion description. }

\textsl{\textbf{Example}: the first person places both hands on the waist while facing the second.}

\textsl{\textbf{Answer}: \{person1: the first person places both hands on the waist while facing the second, person2: the second person also places both hands while facing the other person.\}}
\end{enumerate}
\end{mdframed}

\textbf{Individual task learning.}
Given the fine-grained text prompts $T_t^1$ and $T_t^2$ tailored for each individual, we follow prior works \cite{zhang2022motiondiffuse,zhang2024finemogen,tevet2023human} and utilize the CLIP \cite{radford2021learning} text encoder to extract text encoding features $F_t^1 \in \mathbb{R}^{N \times L} $ and $F_t^2 \in \mathbb{R}^{N_w \times L} $, where $N_w$ represents the number of words and  $L$ is the dimension of word vector. The objective during the self-learning phase is to learn motion encodings that align with individual text prompts. Inspired by \cite{zhang2023remodiffuse} and \cite{shen2021efficient}, we integrate individual motion sequences with individual text prompts to learn a reference sequence through mixed attention. Specifically, we employ the mixed attention on person $i$ motion feature $\mathrm{X}_{mi} \in \mathbb{R}^{S \times D}$ and its individual prompts features $F_t^i$ to capture associated knowledge, which the detail is as: 
\begin{equation}
\mathrm{Query}^i = Q_{mi} \mathrm{X}_{mi},\
\mathrm{Key}^i = [K_{mi} \mathrm{X}_{mi}; K^{i}_{t} F_t^i] ,\
\mathrm{Value}^i = [V_{mi} \mathrm{X}_{mi}; V^{i}_{t} F_t^i],
\label{QKV}
\end{equation}
where $\mathrm{Q}_{m}$, $\mathrm{K}_{m}$, $\mathrm{K}_{t}$, $\mathrm{V}_{m}$, and $\mathrm{V}_{t}$ denote trainable matrices.
\begin{equation}
\mathrm{G^i} = \operatorname{softmax} (\mathrm{Key^i})\mathrm{Value^i},\quad
\mathrm{Y^i} = \operatorname{softmax} (\mathrm{Query^i})\mathrm{G^i}.  
\label{GY}
\end{equation}
By refining encodings through extracting the relationships between text and motion representations within $G^i$, our model progressively incorporates fine-grained semantics across two different modalities. Finally, residualizing the motion features obtains the $\mathrm{X}_{m1}^{s1}$ and $\mathrm{X}_{m2}^{s1}$ as the final output of the first stage.
\begin{equation}
\mathrm{X}_{m1}^{s1} =\mathrm{X}_{m1}+\mathrm{Y^1},\quad
\mathrm{X}_{m2}^{s1} =\mathrm{X}_{m2}+\mathrm{Y^2}.
\label{concat_and_res}
\end{equation}

\subsection{Adaptive adjustment stage} \label{aas}

Previous work \cite{cai2023digital,liang2024intergen} frequently treated interaction as a static and fixed process and ignored the effect of distance that often failed to capture fine-grained interaction details, which is essential for guiding precise dual-human motion generation. 

In real-world scenarios of human interaction, as the distance between two individuals increases, the emphasis shifts towards the quality of their own motion generation, whereas when the distance decreases, the focus heightens on the motions of the interaction partner. Drawing inspiration from this, we designed the adaptive adjustment stage to model the motion interaction at the inter-individual level, which includes the process of human-human interaction distance prediction and interaction-aware graph reasoning.

\textbf{Interaction distance prediction.}
To dynamically adjust the attention based on the interaction distance between individuals, our first objective is to accurately determine this distance across various stages. To achieve this, we developed an Interaction Distance Predictor. Assuming we segment the dynamic interaction process into $K$ distinct segments, inputting an overall text prompt $T_g$ to derive a series of interaction distances $D=\{D_1, D_2, ..., D_K\}$ corresponding to different segments.
In detail, upon receiving the text prompt $T_g$, we initially employ the CLIP \cite{radford2021learning} text encoder to generate the overall text encoding feature $F_g \in \mathbb{R}^{N \times L} $. Subsequently, this feature is translated into predicting interaction distance weight distribution for each segment: 
\begin{equation}
\mathrm{Pre_{dis}} = \operatorname{softmax} (\operatorname{MLP}(F_g)),
\end{equation}
where $\mathrm{Pre_{dis}} \in \mathbb{R}^{K} $. Concurrently, we compute the ground-truth motion's interaction distance weight value, using it as the supervisory signal. For segment $i$ expressed as:
\begin{equation} 
\mathrm{Gt_{dis}^{i}}= \frac{K}{SR} \sum_{l=\frac{(i-1)S}{K} }^{\frac{iS}{K}} \sum_{j=1}^{R} \operatorname{F_{dis}} (\mathrm{gt^{p1}_{jl}} - \mathrm{gt^{p2}_{jl}}),\
\mathrm{F_{dis}}= \sqrt{(x^{p1}-x^{p2})^2 + (y^{p1}-y^{p2})^2 + (z^{p1}-z^{p2})^2}.
\end{equation}
where $\mathrm{gt}$ denotes the 3D joint coordinates of ground-truth motion, $R$ is the joint number, $S$ refers to the motion sequence length, $\frac{(i-1)S}{K}$ to $ \frac{iS}{K}$ denotes the interval length after dividing into $K$ segments, $x^{p1},y^{p1},z^{p1}$ represent the three-dimensional coordinates of the $gt^{p1}$, and $x^{p2},y^{p2},z^{p2}$ represent the three-dimensional coordinates of the $gt^{p2}$. Thus, the ground-truth interaction distance weight distribution is represented as:
\begin{equation} 
\mathrm{Gt_{dis}} = 1 - \operatorname{softmax}\{\mathrm{Gt_{dis}^{1}},\mathrm{Gt_{dis}^{2}},...,\mathrm{Gt_{dis}^{K}}\},
\end{equation}
where $\mathrm{Gt_{dis}} \in \mathbb{R}^{K} $.Consequently, using cross-entropy loss $\mathcal{L}_{2}$ optimizes the distributions between ground truth $\mathrm{Gt_{dis}}$ and predicted $\mathrm{Pre_{dis}}$. 

Overall, our total training loss function is the sum of the motion reconstruction item and the interaction distance optimization item: $\mathcal{L}=\mathcal{L}_{1} + \lambda \mathcal{L}_{2}$, where $\lambda$ is the hyperparameter.

\textbf{Interaction-aware graph reasoning.} 
Modeling interaction distance between individuals effectively captures the dynamic relationship, clarifies each individual's tasks, and mitigates coarse interactions. Specifically, at greater distances, individuals should focus more on their own actions, while at closer distances they must attend to each other's movements to interact effectively. Therefore, we propose interaction-aware graph reasoning, allowing the inference to consider varying interaction intensities based on relative distance between persons to obtain enhanced motion features.

Firstly, we utilize the motion features of two individuals as the initial node features to construct a fully connected graph $G$ representing their interactions. To dynamically model the impact of distance on these interactions, we employ a learnable interaction matrix to adjust the graph topology. This learnable interaction matrix is defined as: $\widetilde{A} = \mathrm{W}_{inter} \mathrm{A}$, where $\mathrm{A} \in \mathbb{R}^{2S \times 2S}$ is the learnable adjacency matrix and $\mathrm{W}_{inter} \in \mathbb{R}^{2S \times 2S}$ is the interaction weight matrix. $\mathrm{W}_{inter}$ is a block matrix with submatrices $I_{S \times S}$ and $w_{S \times S}$, as illustrated in Figure \ref{pipeline} b). Specifically, $I_{S \times S}$ models the motion feature learning within an individual. $w_{S \times S}$ is computed by the product of the coefficients at each segment as the interaction weights (e.g., $0.04 = 0.2 \times 0.2$, $0.16 = 0.2 \times 0.8$, etc.). Finally, based on this interaction graph, a standard graph neural network \cite{scarselli2008graph} module is implemented to learn the motion features that embed the interaction relationships. 

Formally, for person $i$, we concatenate the motion features of the other person along the temporal dimension, yielding $\mathrm{X}_{mi}^{\prime} \in \mathbb{R}^{2S \times D}$, and perform graph reasoning, which is formulated as:
\begin{equation}
\mathrm{G}_{mi} = \sigma (\widetilde{A} X_{mi}^{\prime}),
\label{graph}
\end{equation}
where $\sigma$ denotes the activation function. Finally, we residualize the output of graph reasoning with the output of the previous stage to obtain $\mathrm{X}_{m1}^{s2} $and $\mathrm{X}_{m2}^{s2}$ as the output of the second stage.
\begin{equation}
\mathrm{X}_{mi}^{s2} = \lambda^{s2}_{mi} \mathrm{G}_{mi} +\mathrm{X}_{mi}^{s1}.
\end{equation}
where $\lambda^{s2}_{mi}$ is a hyperparameter. 

\subsection{Teacher-guided refinement stage} \label{tgrs}
In human interaction, each person refines their own motion based on interaction behaviors with others. Thus, we take overall text features as the teacher and propose a teacher-guided refinement stage to refine motion features at overall level. Prior methods only rely on cross-modal attention between text and motion for context, which often fails to convey prompt details and partner interactions, resulting in coarse interactions. These methods also tend to neglect the critical content embedded within motion data. To address this, we propose a teacher-guided refinement stage. During this stage, we initially highlight key content to optimize the features, and then through text-motion interactive learning to further refine the dual-human motion features, as depicted in Figure \ref{pipeline} c).

\textbf{Highlighting key content.}
To emphasize critical keyframes within motion, we focus on key regions in frame channels using computed attention maps between text and motion. Specifically, given motion features of each person, $\mathrm{X}_{m1}^{s2} \in \mathbb{R}^{S \times D}$, $\mathrm{X}_{m2}^{s2} \in \mathbb{R}^{S \times D}$, and overall text feature $F_g \in \mathbb{R}^{N \times L}$, we first employ a linear transformation to get sentence-level text features $F_s \in \mathbb{R}^{1 \times D}$. To spotlight essential frame sequences for Person 1, we enhance it with the text modality, leading to the $text$-$motion_1$ attention map $\mathrm{M}^1_{tm}$ which highlights several the most sentence-relevant keyframes. Similarly, we derive the $text$-$motion_2$ attention map $\mathrm{M}^2_{tm}$. This process can be formulated as:
\begin{equation}
\mathrm{M^1_{tm}}= \operatorname{MLP} (\operatorname{F}_{LN}(\mathrm{X}_{m1}^{s2}) \operatorname{F}_{LN}(F_s)^{\mathsf{T}}),\quad
\mathrm{M^2_{tm}}= \operatorname{MLP} (\operatorname{F}_{LN}(\mathrm{X}_{m2}^{s2}) \operatorname{F}_{LN}(F_s)^{\mathsf{T}}),
\end{equation}
where $\mathrm{M^1_{tm}},\mathrm{M^2_{tm}} \in \mathbb{R}^{S \times D},$ and $\operatorname{F}_{LN}$ represents the LayerNorm function. Finally, we obtain highlighted  key content through residual connections: 
\begin{equation}
\mathrm{X}_{m1}^{s2} = \mathrm{X}_{m1}^{s2} + \lambda_1^{s3} \mathrm{M^1_{tm}},\quad
\mathrm{X}_{m2}^{s2} = \mathrm{X}_{m2}^{s2} + \lambda_2^{s3} \mathrm{M^2_{tm}}.
\end{equation}
where $\lambda_1^{s3}$ and $\lambda_2^{s3}$ are hyperparameters. 

\textbf{Text-motion interactive learning.}
To enhance comprehension of text-motion content, we leverage text, motion, and interaction features through the mixed attention mechanisms mentioned in equation \ref{QKV} and \ref{GY}. In this part, recognizing the character of dual-human motion, we introduce interaction features to serve as additional motion references. More precisely, taking motion1 $\mathrm{X}_{m1}^{s2}$ as an example,
we modify equation \ref{QKV} to incorporate the interaction features by:
\begin{equation}
\mathrm{Query^1} = \mathrm{Q}_{m} \mathrm{X}_{m1}^{s2},\
\mathrm{Key^1} = [\mathrm{K}_{m} \mathrm{X}_{m1}^{s2}; \mathrm{K}_{g} \mathrm{F}_{g}; \mathrm{K}_{h} \mathrm{X}_{m2}^{s2}] ,\
\mathrm{Value^1} = [\mathrm{V}_m \mathrm{X}_{m1}^{s2}; \mathrm{V}_{g} \mathrm{F}_{g}; \mathrm{V}_{h} \mathrm{X}_{m2}^{s2}].
\end{equation}
where $\mathrm{Q}_{m}$, $\mathrm{K}_{m}$, $\mathrm{V}_{m}$, $\mathrm{K}_{g}$, $\mathrm{V}_{g}$, $\mathrm{K}_{h}$ and $\mathrm{V}_{h}$ are trainable interaction weight matrices. Subsequently, operating equation \ref{GY}, we obtained refined features, and through residual processes similar in equation \ref{concat_and_res}, we derived the $\mathrm{X}_{m1}^{s3}$ and $\mathrm{X}_{m2}^{s3}$ as the final output.

\section{Experiments}

\subsection{Datasets, metrics and implementation details}

\textbf{Datasets.} InterHuman \cite{liang2024intergen} is a comprehensive 3D human interactive motion dataset featuring diverse actions performed by two individuals. Each action is annotated with 3 detailed natural languages. The dataset classifies actions into two main categories: everyday actions, which capture interactions like passing objects, greetings, communication, and professional actions, which showcase typical human-human interactions such as taekwondo, Latin dance, and boxing. With a total of 7,779 actions spanning 6.56 hours, the InterHuman dataset offers 23,337 distinct descriptions crafted from 5,656 unique words.

Inter-X \cite{xu2024inter} dataset comprises 13,888 pairs of SMPL-X motion sequences, 34,164 text descriptions, and semantic action categories featuring diverse action/reaction patterns. Additionally, it includes relationship data for 59 groups and personality information for 89 volunteers.

\begin{table*}[t]
\centering
\resizebox{1.\linewidth}{!}{ 
\begin{tabular}{lccccccccc}
\toprule[1.25pt]
\multirow{2}{*}{Methods} & \multicolumn{4}{c}{General Evaluation} & \multicolumn{2}{c}{Individual Evaluation} & \multicolumn{1}{c}{Interaction Evaluation} & \multicolumn{2}{c}{User Study Evaluation} \\
\cmidrule(rl){2-5} \cmidrule(rl){6-7} \cmidrule(rl){8-8} \cmidrule(rl){9-10}
 & FID$\downarrow$ & MM-Dist$\downarrow$ & Diversity$\rightarrow$ & MModality$\uparrow$ & MPJPE$^{p1}$$\downarrow$ & MPJPE$^{p2}$$\downarrow$ & MPJIE$\rightarrow$ & Quality Score$\uparrow$ & Interaction Score$\uparrow$ \\
\midrule
Real motions & $0.273^{\pm 0.007}$ & $3.755^{\pm 0.008}$ & $7.948^{\pm 0.064}$ & - & - & - & $1.246^{\pm 0.001}$ & - & - \\
\midrule
TEMOS~\cite{petrovich2022temos} 
& $17.37^{\pm 0.043}$ 
& $6.342^{\pm 0.015}$ 
& $6.939^{\pm 0.071}$ 
& $0.535^{\pm 0.014}$ 
& - & - & - & - & - \\
T2M~\cite{guo2022generating} 
& $13.76^{\pm 0.072}$ 
& $5.731^{\pm 0.013}$ 
& $7.046^{\pm 0.022}$ 
& $1.387^{\pm 0.076}$  
& - & - & - & - & - \\
MDM~\cite{tevet2023human} 
& $9.167^{\pm 0.056}$ 
& $7.125^{\pm 0.018}$ 
& $7.602^{\pm 0.045}$ 
& $\bm{2.355^{\pm 0.080}}$  
& - & - & - & - & - \\
ComMDM~\cite{shafir2023human} 
& $7.069^{\pm 0.054}$ 
& $6.212^{\pm 0.021}$ 
& $7.244^{\pm 0.038}$ 
& $1.822^{\pm 0.052}$  
& - & - & - & - & - \\
\midrule
in2IN~\cite{Ruiz-Ponce_2024_CVPR} 
& $\underline{5.177^{\pm 0.103}}$ 
& $\underline{3.790^{\pm 0.002}}$ 
& $\underline{7.940^{\pm 0.030}}$ 
& $1.061^{\pm 0.038}$  
& $1.061^{\pm 0.008}$  
& $1.063^{\pm 0.007}$  
& $1.088^{\pm 0.009}$  
& - & -  \\
MotionDiffuse~\cite{zhang2022motiondiffuse} 
& $12.66^{\pm 0.083}$ 
& $3.805^{\pm 0.001}$ 
& $7.639^{\pm 0.035}$ 
& $1.176^{\pm 0.027}$  
& $1.123^{\pm 0.006}$ 
& $1.158^{\pm 0.018}$ 
& $1.023^{\pm 0.005}$ 
& $3.125^{\pm 0.251}$ 
& $3.273^{\pm 0.282}$  \\
ReMoDiffuse~\cite{zhang2023remodiffuse} 
& $6.366^{\pm 0.102}$ 
& $3.802^{\pm 0.001}$ 
& $7.956^{\pm 0.030}$ 
& $1.226^{\pm 0.044}$ 
& $1.074^{\pm 0.003}$ 
& $1.071^{\pm 0.008}$ 
& $1.041^{\pm 0.006}$ 
& $2.972^{\pm 0.307}$ 
& $3.189^{\pm 0.229}$  \\
InterGen~\cite{liang2024intergen} 
& $5.918^{\pm 0.079}$ 
& $5.108^{\pm 0.014}$ 
& $7.387^{\pm 0.029}$ 
& $\underline{2.141^{\pm 0.063}}$  
& $1.066^{\pm 0.005}$ 
& $1.068^{\pm 0.007}$ 
& $1.084^{\pm 0.010}$ 
& $3.328^{\pm 0.343}$ 
& $3.402^{\pm 0.308}$  \\
MoMat-MoGen~\cite{cai2023digital} 
& $5.674^{\pm 0.085}$ 
& $\underline{3.790^{\pm 0.001}}$ 
& $8.021^{\pm 0.035}$ 
& $1.295^{\pm 0.023}$ 
& $\underline{1.060^{\pm 0.013}}$ 
& $\underline{1.062^{\pm 0.008}}$ 
& $\underline{1.091^{\pm 0.006}}$ 
& $3.284^{\pm 0.348}$ 
& $3.341^{\pm 0.299}$  \\
\midrule
\textbf{Ours (FineDual)} 
& $\bm{4.466^{\pm 0.007}}$ 
& $\bm{3.781^{\pm 0.002}}$ 
& $\bm{7.942^{\pm 0.067}}$ 
& $1.303^{\pm 0.096}$
& $\bm{1.015^{\pm 0.009}}$ 
& $\bm{1.021^{\pm 0.007}}$ 
& $\bm{1.125^{\pm 0.010}}$ 
& $\bm{3.753^{\pm 0.184}}$ 
& $\bm{3.881^{\pm 0.207}}$ \\
\bottomrule[1.25pt]
\end{tabular}}
\caption{\textbf{Comparisons to current state-of-the-art methods on the InterHuman \cite{liang2024intergen} test set.} ``$\uparrow$'' denotes that higher is better. ``$\downarrow$'' denotes that lower is better. ``$\rightarrow$'' denotes that results are better if the metric is closer to the real motion. 
We repeat all the evaluations 20 times and report the average with a 95\% confidence interval. \textbf{Bold} and \underline{underlined} indicate the best and second-best results, respectively.}
\label{interhuman_main}
\end{table*}

\begin{table*}[t]
\centering
\resizebox{1.\linewidth}{!}{ 
\begin{tabular}{lcccccccc}
\toprule[1.25pt]
\multirow{2}{*}{Methods} & \multicolumn{3}{c}{R Precision $\uparrow$} & \multirow{2}{*}{FID $\downarrow$} & \multirow{2}{*}{MM Dist $\downarrow$} & \multirow{2}{*}{Diversity $\rightarrow$} & \multirow{2}{*}{MModality $\uparrow$} \\
\cmidrule(rl){2-4}
 & Top 1 & Top 2 & Top 3 & & & & \\
\midrule
Real motions 
& $0.429^{\pm 0.004}$ 
& $0.626^{\pm 0.003}$ 
& $0.736^{\pm 0.003}$ 
& $0.002^{\pm 0.002}$ 
& $0.536^{\pm 0.013}$ 
& $9.734^{\pm 0.078}$ 
& - \\
\midrule
TEMOS~\cite{petrovich2022temos} 
& $0.092^{\pm 0.003}$ 
& $0.171^{\pm 0.003}$ 
& $0.238^{\pm 0.002}$ 
& $29.52^{\pm 0.069}$ 
& $\underline{6.867^{\pm 0.013}}$ 
& $4.738^{\pm 0.078}$ 
& $0.672^{\pm 0.041}$ \\
MDM~\cite{tevet2023human} 
& $0.203^{\pm 0.009}$ 
& $0.329^{\pm 0.007}$ 
& $0.426^{\pm 0.005}$ 
& $23.70^{\pm 0.056}$ 
& $9.548^{\pm 0.014}$ 
& $5.856^{\pm 0.077}$ 
& $\underline{3.490^{\pm 0.061}}$ \\
ComMDM~\cite{shafir2023human} 
& $0.090^{\pm 0.002}$ 
& $0.165^{\pm 0.004}$ 
& $0.236^{\pm 0.004}$ 
& $29.26^{\pm 0.066}$ 
& $6.870^{\pm 0.017}$ 
& $4.734^{\pm 0.067}$ 
& $0.771^{\pm 0.053}$ \\

T2M~\cite{guo2022generating} 
& $0.184^{\pm 0.010}$ 
& $0.298^{\pm 0.006}$ 
& $0.396^{\pm 0.005}$ 
& $5.481^{\pm 0.382}$ 
& $9.576^{\pm 0.006}$ 
& $5.771^{\pm 0.151}$ 
& $2.761^{\pm 0.042}$ \\
InterGen~\cite{liang2024intergen} 
& $\underline{0.207^{\pm 0.004}}$ 
& $\underline{0.335^{\pm 0.005}}$ 
& $\underline{0.429^{\pm 0.005}}$ 
& $\underline{5.207^{\pm 0.216}}$ 
& $9.580^{\pm 0.011}$ 
& $\underline{7.788^{\pm 0.208}}$ 
& $\bm{3.686^{\pm 0.052}}$ \\
\midrule
\textbf{Ours (FineDual)}
& $\bm{0.406^{\pm 0.006}}$ 
& $\bm{0.599^{\pm 0.007}}$ 
& $\bm{0.706^{\pm 0.005}}$ 
& $\bm{0.371^{\pm 0.013}}$ 
& $\bm{3.702^{\pm 0.006}}$ 
& $\bm{9.132^{\pm 0.112}}$ 
& $1.781^{\pm 0.149}$ \\
\bottomrule[1.25pt]
\end{tabular}}
\caption{\textbf{Comparisons to current state-of-the-art methods on the InterX \cite{xu2024inter} test set.}}
\label{interx_main}
\end{table*}

\begin{table*}[t]
\begin{minipage}[t]{0.54\textwidth}
\centering
\resizebox{1.\linewidth}{!}{%
\begin{tabular}{ccc|cccc}
\toprule[1.1pt]
Self- & Adaptive & Teacher-Guided & \multirow{2}{*}{FID $\downarrow$} & \multirow{2}{*}{MPJPE$^{p1}$ $\downarrow$} & \multirow{2}{*}{MPJPE$^{p2}$ $\downarrow$} & \multirow{2}{*}{MPJIE $\rightarrow$} \\
Learning Stage & Adjustment Stage & Refinement Stage \\ \midrule
 & \checkmark & \checkmark & $4.752^{\pm 0.008}$ & $1.322^{\pm 0.008}$ & $1.367^{\pm 0.005}$ & $1.017^{\pm 0.009}$ \\
\checkmark & & \checkmark & $4.862^{\pm 0.007}$ & $1.215^{\pm 0.014}$ & $1.253^{\pm 0.010}$ & $0.972^{\pm 0.009}$ \\
\checkmark & \checkmark & & $7.913^{\pm 0.010}$ & $1.615^{\pm 0.011}$ & $1.672^{\pm 0.009}$ & $0.926^{\pm 0.007}$ \\
\checkmark & \checkmark & \checkmark & $\bm{4.466^{\pm 0.007}}$ & $\bm{1.015^{\pm 0.009}}$ & $\bm{1.021^{\pm 0.007}}$ & $\bm{1.125^{\pm 0.010}}$ \\
\bottomrule[1.1pt]
\end{tabular}%
}
\caption{\textbf{Ablation study on the InterHuman \cite{liang2024intergen} test set.}}
\label{ablation1}
\end{minipage}
\quad
\hfill
\begin{minipage}[t]{0.43\textwidth}
\centering
\resizebox{1.\linewidth}{!}{%
\begin{tabular}{ccc|cccc}
\toprule[1.1pt]
Individual & Graph & Multimodal & \multirow{2}{*}{FID $\downarrow$} & \multirow{2}{*}{MPJPE$^{p1}$ $\downarrow$} & \multirow{2}{*}{MPJPE$^{p2}$ $\downarrow$} & \multirow{2}{*}{MPJIE $\rightarrow$} \\
Layer & Layer & Layer \\ \midrule
1 & 1 & 3 & $4.835^{\pm 0.004}$ & $1.286^{\pm 0.007}$ & $1.325^{\pm 0.006}$ & $0.887^{\pm 0.008}$ \\
1 & 1 & 5 & $4.958^{\pm 0.005}$ & $1.294^{\pm 0.009}$ & $1.322^{\pm 0.005}$ & $0.895^{\pm 0.009}$ \\
1 & 2 & 3 & $4.562^{\pm 0.007}$ & $1.065^{\pm 0.009}$ & $1.047^{\pm 0.007}$ & $1.115^{\pm 0.010}$ \\
1 & 3 & 3 & $4.512^{\pm 0.006}$ & $1.158^{\pm 0.008}$ & $1.183^{\pm 0.007}$ & $1.042^{\pm 0.007}$ \\
2 & 2 & 3 & $\bm{4.466^{\pm 0.007}}$ & $\bm{1.015^{\pm 0.009}}$ & $\bm{1.021^{\pm 0.007}}$ & $\bm{1.125^{\pm 0.010}}$ \\
3 & 2 & 3 & $4.681^{\pm 0.003}$ & $1.076^{\pm 0.010}$ & $1.089^{\pm 0.009}$ & $1.049^{\pm 0.011}$ \\
\bottomrule[1.1pt]
\end{tabular}%
}
\caption{\textbf{Ablation study on the InterHuman \cite{liang2024intergen} test set.}}
\label{layers}
\end{minipage}
\hfill
\end{table*}

\textbf{Metrics.}
To evaluate dual-human motion generation, we first employ general metrics to assess the quality of the generation. We followed \cite{guo2022generating,liang2024intergen} general metrics such as \textbf{R Precision}, \textbf{FID}, \textbf{MM-Dist}, \textbf{Diversity}, and \textbf{Multimodality}. 
\textbf{R Precision} is calculated by evaluating the Top-1/2/3 matching accuracy between the text and motion. \textbf{FID} measures the similarity between the feature distributions extracted from the generated motions and the ground truth motions. \textbf{MM-Dist} computes the average Euclidean distance between the feature of generated motions and the text prompt feature. \textbf{Diversity} evaluates the dissimilarity among all generated motions across all descriptions. \textbf{Multimodality} measures the average variance of generated motions for a given text prompt.

Besides, we introduce several metrics to evaluate the quality of dual-human fine-grained generation. (1) \textbf{Mean Per Joint Position Error (MPJPE)} is used to evaluate the motion quality of each individual. For person 1 (p1), computing the mean difference between all joints of the model generated ($\textit{pre}$) and their corresponding ground truth motions ($\textit{gt}$): 
$\operatorname{MPJPE}^{p1}= \frac{1}{TN_j} \sum_{i=1}^{T} \sum_{j=1}^{N_j} \operatorname{F_{dis}} (pre^{p1}_{ij} - gt^{p1}_{ij})$, where $T$ denotes the motion sequence and $N_j$ is the joint number. 
(2) \textbf{Mean Per Joint Interaction Error (MPJIE)} is used to evaluate the interaction quality between two individuals. For a group of two people, the MPJIE is outlined as follows:
$\operatorname{MPJIE}= \frac{1}{TN_j}\sum_{i=1}^{T} \sum_{j=1}^{N_j} \operatorname{F_{dis}} (pre^{p_1}_{ij} - pre^{p_2}_{ij}) $, where $p_1$ and $p_2$ represent person 1 and person 2, respectively.
MPJIE provides insight into the relative positioning of individuals, which is an important aspect of physical plausibility and spatial consistency, and serves as a complementary quantitative metric for measuring spatial alignment between interacting humans.
(3) We conducted a user study to assess the generated results in terms of overall \textbf{Quality Score} and \textbf{Interaction Score}, which represent users' evaluations of the overall motion quality and interaction quality of the dual-human motion, respectively.
Given the generated motions and text prompts, participants were prompted to assess the motions of quality score and interactive score, using a 1 to 5 rating scale. This study engaged 30 participants, each evaluating 10 motions for every method.

\textbf{Implementation details.}
Regarding the motion encoder, we employ a 2-layer transformer in the self-learning stage and a 3-layer transformer in the teacher-guided refinement stage with a latent dimension of 512 for each person. As for the text encoder, a frozen text encoder from CLIP ViT-B/32 is utilized, complemented by two additional transformer encoder layers. Regarding some hyperparameters, $D_K$ is set to 3, $\lambda^{s2}$ and $\lambda^{s3}$ are set to 0.1, and the guidance scale $s$ is set to 1.8. In terms of the diffusion model, the variances $\beta_t$ are predefined to linearly spread from 0.0001 to 0.02, and the total number of noising steps is set at T = 1000. We use the Adam optimizer to train the model with an initial learning rate of 0.0002, gradually decreasing to 0.00002 through a cosine learning rate scheduler. The training process is conducted on 2 NVIDIA GeForce RTX 3090, with a batch size of 80 on a single GPU.

For pose representation, we follow Liang \textit{et al.} \cite{liang2024intergen}. The pose states contain four different parts: ($j^{p}_{g}, j^{v}_{g},j^{r},c^{f}$). Here $j^{p}_{g} \in \mathbb{R}^{3N_{j}}$ are the global joint positions. $j^{p}_{g} \in \mathbb{R}^{3N_{j}}$ are velocities. $j^{r} \in \mathbb{R}^{6N_{j}}$ are 6D representation of local rotations and $c^{f} \in \mathbb{R}^{4}$ are binary foot-ground contact features.

\subsection{Evaluation of dual-human motion generation}

\paragraph{General evaluation} 
The results in Table \ref{interhuman_main} compare FineDual against state-of-the-art (SOTA) methods including TEMOS \cite{petrovich2022temos}, T2M \cite{guo2022generating}, MDM \cite{tevet2023human}, ComMDM \cite{shafir2023human}, in2IN \cite{Ruiz-Ponce_2024_CVPR}, MotionDiffuse \cite{zhang2022motiondiffuse}, ReMoDiffuse \cite{zhang2023remodiffuse}, InterGen \cite{liang2024intergen}, and MoMat-MoGen \cite{cai2023digital}. FineDual achieves significantly better scores than all methods in FID and MM-Dist. For example, compared to the current sota method MoMat-MoGen \cite{cai2023digital}, FineDual has achieved a reduction of 21.29\% in the FID metric, which indicates its ability to generate high-fidelity motions closely matching text semantics. For diversity metrics (Diversity and MModality), FineDual obtains marginally lower scores in MModality, likely owing to its emphasis on hierarchical refinement to adhere precisely to individual and overall prompts. While this constrains motion variety slightly, prioritizing accuracy is important, as imprecise alignments diminish the value of diversity. Overall, the results demonstrate that FineDual produces quantitatively improved dual-human motion while reasonably balancing the trade-off between accuracy and diversity.

The results in Table \ref{interx_main} compare FineDual against SOTA methods including TEMOS \cite{petrovich2022temos}, T2M \cite{guo2022generating}, MDM \cite{tevet2023human}, ComMDM \cite{shafir2023human}, and InterGen \cite{liang2024intergen} on the Inter-X \cite{xu2024inter} dataset. FineDual outperforms all existing methods by a significant margin across all evaluation metrics. Notably, it achieves a 64.57\% improvement in R-Top3 accuracy compared to InterGen \cite{liang2024intergen}. Furthermore, FineDual reduces the FID and MM-DIST scores by 4.836 and 5.878, respectively, demonstrating substantial quantitative advancements. These results confirm that our approach effectively captures higher-fidelity human-human interactions. Overall, FineDual sets a new state-of-the-art performance on the Inter-X \cite{xu2024inter} dataset.

\paragraph{Fine-grained evaluation}
Existing general evaluation lacks finesse, only assessing overall dual-human motion quality. This fails to precisely measure individual motion generation and interaction effectiveness. Thus, we perform individual evaluation, interaction evaluation, and user study evaluations against MotionDiffuse \cite{zhang2022motiondiffuse}, in2IN \cite{Ruiz-Ponce_2024_CVPR}, ReMoDiffuse \cite{zhang2023remodiffuse}, InterGen \cite{liang2024intergen} and MoMat-MoGen \cite{cai2023digital} as shown in Table \ref{interhuman_main}. FineDual's specialized self-learning markedly improves the MPJPE metric for individual motion accuracy. Its adaptive adjustment also enhances MPJIE for interaction capture by modeling dynamic relationships. Specifically, compared to in2IN, a method that utilizes both interaction-level and individual-level textual descriptions, our approach demonstrates consistent performance gains: 4.3\% and 3.9\% improvements on the individual evaluation ($\text{MPJPE}^{p1}$ and $\text{MPJPE}^{p2}$), along with a 3.3\% enhancement on the interaction evaluation (MPJIE).
User study further confirms that FineDual surpasses competitors in motion and interaction quality. For comparison with InterGen \cite{liang2024intergen}, our method demonstrates a notable improvement, with the quality score exceeding that of InterGen by 11.3\%, and the interaction score showing an enhancement of 12.4\%. This fine-grained analysis substantiates FineDual's ability to produce high-fidelity individual motions, demonstrating its excellence in fine-grained motion generation.

\begin{figure*}[t]
    \centering
    \includegraphics[width=\linewidth]{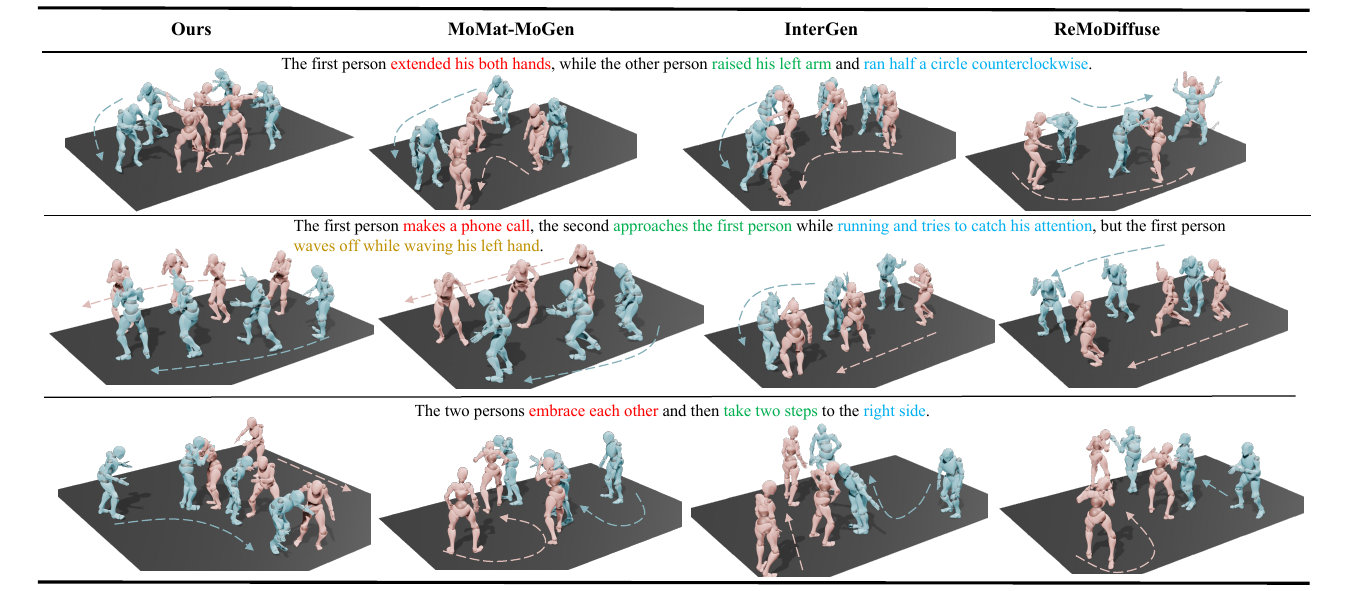}
    \caption{\textbf{Visual results compared with existing methods.} The arrows represent the time axes.}
    \label{compare}
\end{figure*}

\begin{figure*}[t]
    \centering
    \includegraphics[width=\linewidth]{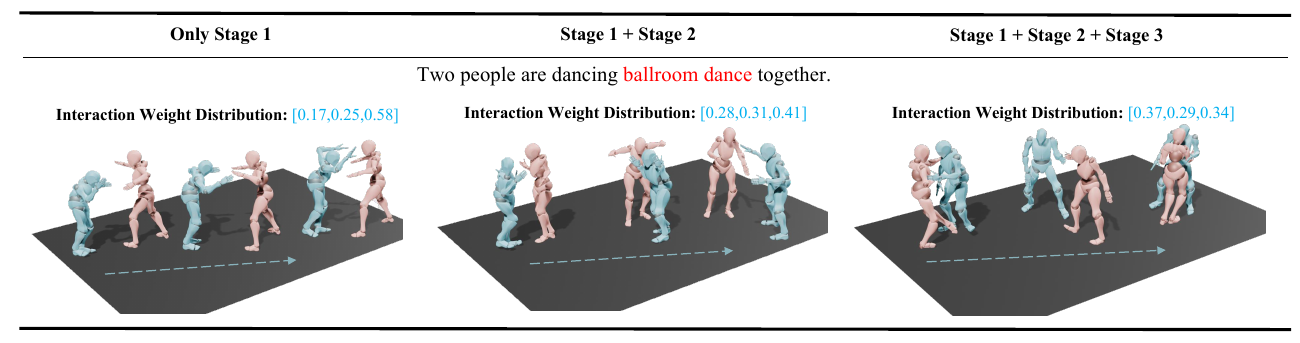}
    \caption{\textbf{Qualitative comparison of different stages.} Stage 1, stage 2, and stage 3 represent Self-Learning Stage, Adaptive Adjustment Stage, and Teacher-Guided Refinement Stage respectively. The arrows represent the time axes.}
    \label{3stage}
\end{figure*}

\paragraph{Qualitative analysis}
Figure \ref{compare} qualitatively compares FineDual against ReMoDiffuse \cite{zhang2023remodiffuse}, InterGen \cite{liang2024intergen}, and MoMat-MoGen \cite{cai2023digital}. While others struggle with these prompts, FineDual demonstrates success. 
ReMoDiffuse \cite{zhang2023remodiffuse} incorrectly implements opposing directions to those specified. InterGen \cite{liang2024intergen} and MoMat-MoGen \cite{cai2023digital} fail to interpret requests to raise the left or right hand as indicated in the language. Additionally, their interactions are not always accurately coordinated, sometimes showing improbable overlap or disconnectedness between individuals. In contrast, FineDual executes motions adhering closely to linguistic specifications for behaviors like specific hand movements. Its dedicated modeling for individual actions and relationships is evident in the smooth, coordinating dual-human motion generation. This analysis illustrates FineDual's effectiveness in generating the detailed motions and precise dynamics described in natural language inputs.

\paragraph{Ablation study}

Table \ref{ablation1} presents ablation results assessing the three stages of our method.
When the self-learning stage is excluded, the model's performance declines across FID, MPJPE, and MPJIE metrics. Specifically, MPJPE deteriorates by 23.2\% and 25.3\% for the two subjects, respectively. 
Notably, since LLMs decompose prompts into single-person texts, facilitating independent learning for each individual, the absence of this phase significantly degrades MPJPE. These results highlight the advantages of specialized individual motion modeling.
Removing the adaptive adjustment stage also reduces performance across all metrics. More critically, as this phase incorporates interaction distance into graph reasoning to capture detailed interpersonal relationships, its absence degrades MPJIE (a 15.7\% reduction), reflecting poorer interaction quality.
When the teacher-guided refinement stage is omitted, there is a substantial drop in the model's performance on all three motion quality metrics, given its role as the core module for learning dual-human text prompts. 
Additionally, without the third phase, the model fails to effectively comprehend dual-human interaction text prompts, significantly degrading the FID metric by 20\% - a crucial metric for assessing dual-human motion generation quality.
The results of the ablation analysis validated the effectiveness of each stage in FineDual. Our full model achieves optimal performance across metrics, validating the importance of these components working synergistically.

\begin{figure*}[t]
    \centering
    \includegraphics[width=\linewidth]{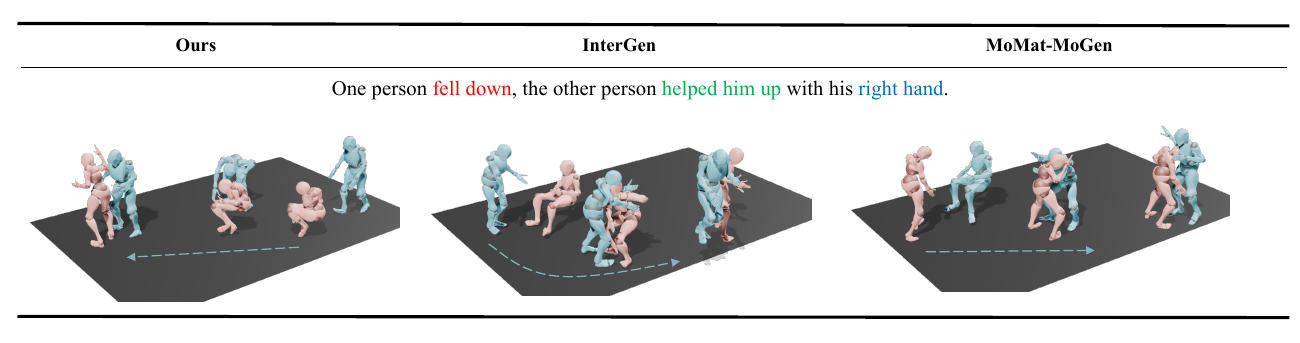}
    \caption{\textbf{Qualitative analysis on the coarse-grained interaction problem.} The arrows represent the time axes.}
    \label{corse-grained}
\end{figure*}
\begin{figure*}[t]
    \centering
    \includegraphics[width=\linewidth]{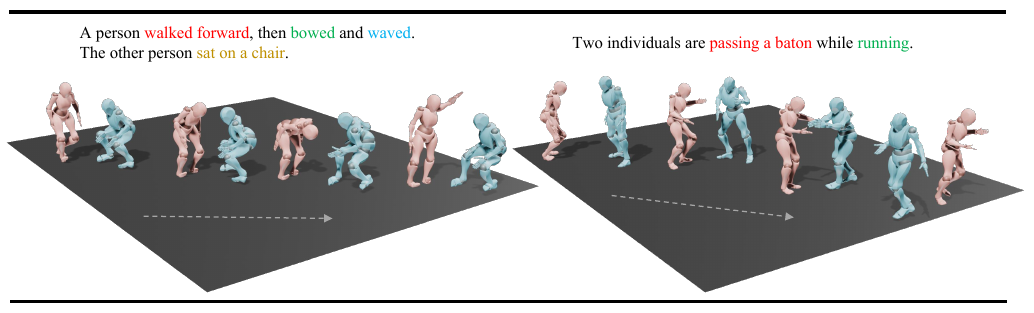}
    \caption{\textbf{More visual examples.} The arrows represent the time axes.}
    \label{morevis_2person}
\end{figure*}
We have conducted additional ablation experiments on the number of layers, as shown in Table \ref{layers}. The experiments analyze different layers in our individual task learning, interaction-aware graph reasoning, and text-motion interactive learning components. The individual layers positively impact MPJPE through enhanced individual representation. However, excess layers can overemphasize individual modeling at the cost of decreased overall motion quality FID. The graph layers help optimize MPJIE by incorporating interaction distance. Meanwhile, the multimodal layers refine cross-modal alignment, significantly boosting FID via balanced learning of text, motion, and interactions. Nevertheless, too many layers only provide marginal gains while increasing computational overhead. Overall, our method strikes the right balance across these factors through careful architectural tuning, simultaneously optimizing individual, interaction, and overall generation metrics.

Moreover, we conducted a visualization comparison across different stages in Figure \ref{3stage}. Initially, when only the first stage—the self-learning stage—is implemented, the motion captures merely the basic essence of ballroom dance, but the quality of the motion is unsatisfactory. Adding the second stage, the adaptive adjustment stage substantially enhances the quality of the interactions between people, but the interaction between individuals is still insufficient. When all three stages are integrated, the resulting motion sequences not only showcase rich semantic content but also demonstrate more accurate and detailed motion interactions. 
Meanwhile, we present visualizations of interaction distance weight distribution across different stages in Figure \ref{3stage}. The results show that when only the initial stage (Self-Learning Stage) is implemented, the distribution of interaction weights is suboptimal, skewing disproportionately toward the latter part of the motion sequence. However, ballroom dancing inherently involves a continuous, intense interaction. By integrating the second (Adaptive Adjustment Stage) and third (Teacher-Guided Refinement Stage) stages, FineDual achieves a more balanced and appropriate allocation of motion interaction weights, reflecting the ongoing dynamics of ballroom interactions.

\paragraph{Qualitative analysis on the coarse-grained interaction problem}
To highlight the issues of coarse-grained interactions in other methods, we conducted experiments depicted in Figure \ref{corse-grained}. The motion sequences generated by the InterGen \cite{liang2024intergen} and Momat-MoGen \cite{cai2023digital} frequently show clipping during human-to-human interactions and display a notable discrepancy in the alignment of generated motions with the intended text semantics. 
Specifically, these method fail to demonstrate the `fell down' action. In contrast, our approach effectively maintains textual semantic consistency by realistically generating motions like `helped him up with his right hand', capturing the nuanced dynamics of human-to-human interactions.

\paragraph{Additional visualization results}
Figure \ref{morevis_2person} provides additional visualizations of dual-human interactions. 
From the first example in the figure, it can be observed that our method demonstrates excellent individual-level motion performance across multiple consecutive actions, including walking, bowing, and waving.
From the second example, our method effectively achieves passing a baton while running, showcasing outstanding interaction quality and motion coordination.
It demonstrates that our method's precision in generating motion sequences from textual prompts and showcasing high-quality human-human interactions. Further comparisons and examples are available in the supplementary video.

\paragraph{Failure cases}
\begin{figure*}[t]
    \centering
    \includegraphics[width=0.8\linewidth]{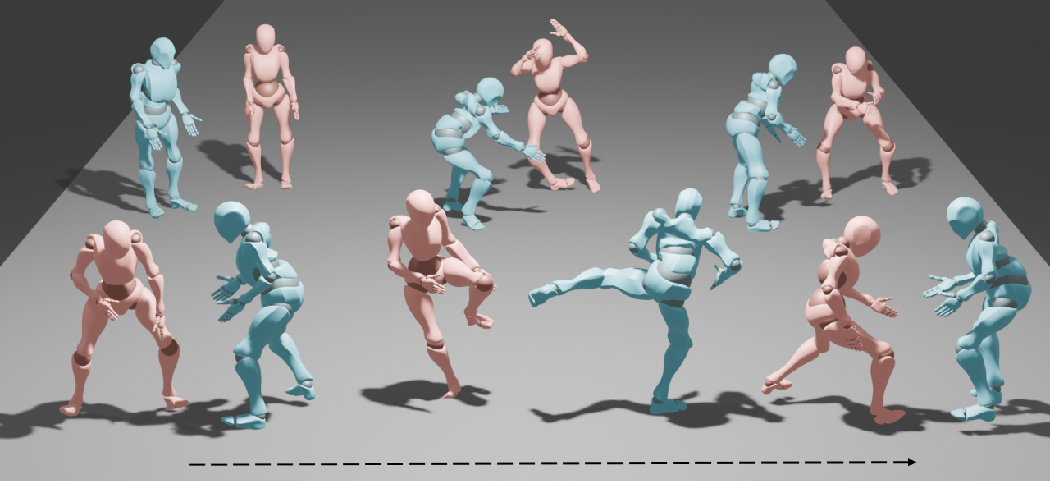}
    \caption{\textbf{Visualization of the failure cases}. Front): Two people are practicing kung fu. Back): Two people are playing a vault game. The arrows represent the time axes.}
    \label{failure}
\end{figure*}
Finally, we present the failure cases in Figure \ref{failure}. While FineDual effectively models fine-grained human interactions, it encounters challenges with unseen texts. For the front text, our approach of personalized task learning leveraged LLMs to decompose ``Kung Fu" into well-understood kicking motions, which have been previously learned. However, the back text, ``vault," was not covered in the dataset, resulting in a lower performance. Future efforts can focus on enhancing the model's generalization capabilities.

\section{Conclusion}
In this paper, we introduce FineDual, a text-driven method for generating fine-grained dual-human interaction motions. FineDual utilizes a dynamic hierarchical interaction framework to iteratively refine human-human interactions through three key stages:
(1) Self-Learning Stage divides the dual-human overall text into individual texts through LLM. (2) Adaptive Adjustment Stage predicts interaction distance by an interaction distance predictor, modeling human interactions dynamically by an interaction-aware graph network. (3) Teacher-Guided Refinement Stage utilizes overall text features as guidance to refine motion features, generating fine-grained dual-human motion. 
Comprehensive quantitative and qualitative evaluations on InterHuman and Inter-X datasets demonstrate that our proposed FineDual outperforms existing approaches, effectively modeling dynamic hierarchical human interaction.

\section{Limitation and future work}
Although our method has achieved some advancements, it still encounters some limitations that merit further investigation. (i) In the adaptive adjustment stage, our current approach requires manual segmentation of person-to-person interactions into several subparts, i.e. setting hyperparameter $D_K$. In the future, the development of a flexible algorithm capable of automatically segmenting these interactions into arbitrary sub-processes suitable for motions could profoundly enhance the understanding of human-human interaction dynamics and provide significant benefits to the broader research community. (ii) Due to the absence of text-driven multi-human datasets, our capacity to generate a diverse range of multi-human motions remains constrained. The creation of a comprehensive, high-quality multi-human interaction dataset represents a valuable avenue for future research. Meanwhile, exploring multi-human generation paradigms is equally crucial, as this may inherently involve temporal efficiency challenges. A potential solution would require first developing a group motion planner to pre-coordinate multi-human interactions, thereby enabling scalable generation for arbitrary numbers of participants.

\Acknowledgements {This work was supported by the Academic Excellence Foundation of BUAA for PhD Students and the National Natural Science Foundation of China (Project Number: 62272019).}

\bibliographystyle{plain}
\bibliography{main.bib}

\begin{thebibliography}{10}

\bibitem{achiam2023gpt}
Josh Achiam, Steven Adler, Sandhini Agarwal, Lama Ahmad, Ilge Akkaya, Florencia~Leoni Aleman, Diogo Almeida, Janko Altenschmidt, Sam Altman, Shyamal Anadkat, et~al.
\newblock Gpt-4 technical report.
\newblock {\em arXiv preprint arXiv:2303.08774}, 2023.

\bibitem{ahuja2019language2pose}
Chaitanya Ahuja and Louis-Philippe Morency.
\newblock Language2pose: Natural language grounded pose forecasting.
\newblock In {\em 2019 International Conference on 3D Vision (3DV)}, pages 719--728. IEEE, 2019.

\bibitem{batz2009recognition}
Thomas Batz, Kym Watson, and Jurgen Beyerer.
\newblock Recognition of dangerous situations within a cooperative group of vehicles.
\newblock In {\em 2009 IEEE Intelligent Vehicles Symposium}, pages 907--912. IEEE, 2009.

\bibitem{bhattacharya2021text2gestures}
Uttaran Bhattacharya, Nicholas Rewkowski, Abhishek Banerjee, Pooja Guhan, Aniket Bera, and Dinesh Manocha.
\newblock Text2gestures: A transformer-based network for generating emotive body gestures for virtual agents.
\newblock In {\em 2021 IEEE virtual reality and 3D user interfaces (VR)}, pages 1--10. IEEE, 2021.

\bibitem{cai2023digital}
Zhongang Cai, Jianping Jiang, Zhongfei Qing, Xinying Guo, Mingyuan Zhang, Zhengyu Lin, Haiyi Mei, Chen Wei, Ruisi Wang, Wanqi Yin, et~al.
\newblock Digital life project: Autonomous 3d characters with social intelligence.
\newblock {\em arXiv preprint arXiv:2312.04547}, 2023.

\bibitem{cervantes2022implicit}
Pablo Cervantes, Yusuke Sekikawa, Ikuro Sato, and Koichi Shinoda.
\newblock Implicit neural representations for variable length human motion generation.
\newblock In {\em European Conference on Computer Vision}, pages 356--372. Springer, 2022.

\bibitem{chen2023executing}
Xin Chen, Biao Jiang, Wen Liu, Zilong Huang, Bin Fu, Tao Chen, and Gang Yu.
\newblock Executing your commands via motion diffusion in latent space.
\newblock In {\em Proceedings of the IEEE/CVF Conference on Computer Vision and Pattern Recognition}, pages 18000--18010, 2023.

\bibitem{guo2023momask}
Chuan Guo, Yuxuan Mu, Muhammad~Gohar Javed, Sen Wang, and Li~Cheng.
\newblock Momask: Generative masked modeling of 3d human motions.
\newblock {\em arXiv preprint arXiv:2312.00063}, 2023.

\bibitem{guo2022generating}
Chuan Guo, Shihao Zou, Xinxin Zuo, Sen Wang, Wei Ji, Xingyu Li, and Li~Cheng.
\newblock Generating diverse and natural 3d human motions from text.
\newblock In {\em CVPR}, pages 5152--5161, 2022.

\bibitem{guo2022tm2t}
Chuan Guo, Xinxin Zuo, Sen Wang, and Li~Cheng.
\newblock Tm2t: Stochastic and tokenized modeling for the reciprocal generation of 3d human motions and texts.
\newblock In {\em European Conference on Computer Vision}, pages 580--597. Springer, 2022.

\bibitem{guo2020action2motion}
Chuan Guo, Xinxin Zuo, Sen Wang, Shihao Zou, Qingyao Sun, Annan Deng, Minglun Gong, and Li~Cheng.
\newblock Action2motion: Conditioned generation of 3d human motions.
\newblock In {\em Proceedings of the 28th ACM International Conference on Multimedia}, pages 2021--2029, 2020.

\bibitem{habibi2018context}
Golnaz Habibi, Nikita Jaipuria, and Jonathan~P How.
\newblock Context-aware pedestrian motion prediction in urban intersections.
\newblock {\em arXiv preprint arXiv:1806.09453}, 2018.

\bibitem{hall1966hidden}
Edmund~T Hall and Edward~T Hall.
\newblock {\em The hidden dimension}, volume 609.
\newblock Anchor, 1966.

\bibitem{ho2022classifier}
Jonathan Ho and Tim Salimans.
\newblock Classifier-free diffusion guidance.
\newblock {\em arXiv preprint arXiv:2207.12598}, 2022.

\bibitem{jin2024act}
Peng Jin, Yang Wu, Yanbo Fan, Zhongqian Sun, Wei Yang, and Li~Yuan.
\newblock Act as you wish: Fine-grained control of motion diffusion model with hierarchical semantic graphs.
\newblock {\em Advances in Neural Information Processing Systems}, 36, 2024.

\bibitem{kao2020temporally}
Hsuan-Kai Kao and Li~Su.
\newblock Temporally guided music-to-body-movement generation.
\newblock In {\em Proceedings of the 28th ACM International Conference on Multimedia}, pages 147--155, 2020.

\bibitem{karunratanakul2023guided}
Korrawe Karunratanakul, Konpat Preechakul, Supasorn Suwajanakorn, and Siyu Tang.
\newblock Guided motion diffusion for controllable human motion synthesis.
\newblock In {\em Proceedings of the IEEE/CVF International Conference on Computer Vision}, pages 2151--2162, 2023.

\bibitem{li2021ai}
Ruilong Li, Shan Yang, David~A Ross, and Angjoo Kanazawa.
\newblock Ai choreographer: Music conditioned 3d dance generation with aist++.
\newblock In {\em Proceedings of the IEEE/CVF International Conference on Computer Vision}, pages 13401--13412, 2021.

\bibitem{liang2024intergen}
Han Liang, Wenqian Zhang, Wenxuan Li, Jingyi Yu, and Lan Xu.
\newblock Intergen: Diffusion-based multi-human motion generation under complex interactions.
\newblock {\em International Journal of Computer Vision}, pages 1--21, 2024.

\bibitem{lucas2022posegpt}
Thomas Lucas, Fabien Baradel, Philippe Weinzaepfel, and Gr{\'e}gory Rogez.
\newblock Posegpt: Quantization-based 3d human motion generation and forecasting.
\newblock In {\em European Conference on Computer Vision}, pages 417--435. Springer, 2022.

\bibitem{petrovich2021action}
Mathis Petrovich, Michael~J Black, and G{\"u}l Varol.
\newblock Action-conditioned 3d human motion synthesis with transformer vae.
\newblock In {\em Proceedings of the IEEE/CVF International Conference on Computer Vision}, pages 10985--10995, 2021.

\bibitem{petrovich2022temos}
Mathis Petrovich, Michael~J Black, and G{\"u}l Varol.
\newblock Temos: Generating diverse human motions from textual descriptions.
\newblock In {\em ECCV}, pages 480--497, 2022.

\bibitem{pinyoanuntapong2023mmm}
Ekkasit Pinyoanuntapong, Pu~Wang, Minwoo Lee, and Chen Chen.
\newblock Mmm: Generative masked motion model.
\newblock {\em arXiv preprint arXiv:2312.03596}, 2023.

\bibitem{radford2021learning}
Alec Radford, Jong~Wook Kim, Chris Hallacy, Aditya Ramesh, Gabriel Goh, Sandhini Agarwal, Girish Sastry, Amanda Askell, Pamela Mishkin, Jack Clark, et~al.
\newblock Learning transferable visual models from natural language supervision.
\newblock In {\em International conference on machine learning}, pages 8748--8763. PMLR, 2021.

\bibitem{ren2020self}
Xuanchi Ren, Haoran Li, Zijian Huang, and Qifeng Chen.
\newblock Self-supervised dance video synthesis conditioned on music.
\newblock In {\em Proceedings of the 28th ACM International Conference on Multimedia}, pages 46--54, 2020.

\bibitem{Ruiz-Ponce_2024_CVPR}
Pablo Ruiz-Ponce, German Barquero, Cristina Palmero, Sergio Escalera, and Jos\'e Garc{\'\i}a-Rodr{\'\i}guez.
\newblock in2in: Leveraging individual information to generate human interactions.
\newblock In {\em Proceedings of the IEEE/CVF Conference on Computer Vision and Pattern Recognition (CVPR) Workshops}, pages 1941--1951, June 2024.

\bibitem{scarselli2008graph}
Franco Scarselli, Marco Gori, Ah~Chung Tsoi, Markus Hagenbuchner, and Gabriele Monfardini.
\newblock The graph neural network model.
\newblock {\em IEEE transactions on neural networks}, 20(1):61--80, 2008.

\bibitem{shafir2023human}
Yoni Shafir, Guy Tevet, Roy Kapon, and Amit~Haim Bermano.
\newblock Human motion diffusion as a generative prior.
\newblock In {\em The Twelfth International Conference on Learning Representations}, 2023.

\bibitem{shen2021efficient}
Zhuoran Shen, Mingyuan Zhang, Haiyu Zhao, Shuai Yi, and Hongsheng Li.
\newblock Efficient attention: Attention with linear complexities.
\newblock In {\em Proceedings of the IEEE/CVF winter conference on applications of computer vision}, pages 3531--3539, 2021.

\bibitem{starke2022deepphase}
Sebastian Starke, Ian Mason, and Taku Komura.
\newblock Deepphase: Periodic autoencoders for learning motion phase manifolds.
\newblock {\em ACM Transactions on Graphics (TOG)}, 41(4):1--13, 2022.

\bibitem{tanaka2023role}
Mikihiro Tanaka and Kent Fujiwara.
\newblock Role-aware interaction generation from textual description.
\newblock In {\em Proceedings of the IEEE/CVF International Conference on Computer Vision}, pages 15999--16009, 2023.

\bibitem{tang2018long}
Yongyi Tang, Lin Ma, Wei Liu, and Wei-Shi Zheng.
\newblock Long-term human motion prediction by modeling motion context and enhancing motion dynamics.
\newblock In {\em Proceedings of the Twenty-Seventh International Joint Conference on Artificial Intelligence}. International Joint Conferences on Artificial Intelligence Organization, 2018.

\bibitem{tevet2023human}
Guy Tevet, Sigal Raab, Brian Gordon, Yoni Shafir, Daniel Cohen-or, and Amit~Haim Bermano.
\newblock Human motion diffusion model.
\newblock In {\em The Eleventh International Conference on Learning Representations}, 2023.

\bibitem{tseng2023edge}
Jonathan Tseng, Rodrigo Castellon, and Karen Liu.
\newblock Edge: Editable dance generation from music.
\newblock In {\em Proceedings of the IEEE/CVF Conference on Computer Vision and Pattern Recognition}, pages 448--458, 2023.

\bibitem{wan2023tlcontrol}
Weilin Wan, Zhiyang Dou, Taku Komura, Wenping Wang, Dinesh Jayaraman, and Lingjie Liu.
\newblock Tlcontrol: Trajectory and language control for human motion synthesis.
\newblock {\em arXiv preprint arXiv:2311.17135}, 2023.

\bibitem{wang2025most}
Yin Wang, Zhiying Leng, Frederick~WB Li, Xiaohui Liang, et~al.
\newblock Most: Motion diffusion model for rare text via temporal clip banzhaf interaction.
\newblock {\em IEEE Transactions on Visualization and Computer Graphics}, 2025.

\bibitem{wang2023fg}
Yin Wang, Zhiying Leng, Frederick~WB Li, Shun-Cheng Wu, and Xiaohui Liang.
\newblock Fg-t2m: Fine-grained text-driven human motion generation via diffusion model.
\newblock In {\em Proceedings of the IEEE/CVF International Conference on Computer Vision}, pages 22035--22044, 2023.

\bibitem{wang2025fg}
Yin Wang, Mu~Li, Jiapeng Liu, Zhiying Leng, Frederick~WB Li, Ziyao Zhang, and Xiaohui Liang.
\newblock Fg-t2m++: Llms-augmented fine-grained text driven human motion generation.
\newblock {\em International Journal of Computer Vision}, pages 1--17, 2025.

\bibitem{xu2024inter}
Liang Xu, Xintao Lv, Yichao Yan, Xin Jin, Shuwen Wu, Congsheng Xu, Yifan Liu, Yizhou Zhou, Fengyun Rao, Xingdong Sheng, et~al.
\newblock Inter-x: Towards versatile human-human interaction analysis.
\newblock In {\em Proceedings of the IEEE/CVF Conference on Computer Vision and Pattern Recognition}, pages 22260--22271, 2024.

\bibitem{zhang2023generating}
Jianrong Zhang, Yangsong Zhang, Xiaodong Cun, Yong Zhang, Hongwei Zhao, Hongtao Lu, Xi~Shen, and Ying Shan.
\newblock Generating human motion from textual descriptions with discrete representations.
\newblock In {\em Proceedings of the IEEE/CVF Conference on Computer Vision and Pattern Recognition}, pages 14730--14740, 2023.

\bibitem{zhang2022motiondiffuse}
Mingyuan Zhang, Zhongang Cai, Liang Pan, Fangzhou Hong, Xinying Guo, Lei Yang, and Ziwei Liu.
\newblock Motiondiffuse: Text-driven human motion generation with diffusion model.
\newblock {\em IEEE Transactions on Pattern Analysis and Machine Intelligence}, pages 1--15, 2024.

\bibitem{zhang2023remodiffuse}
Mingyuan Zhang, Xinying Guo, Liang Pan, Zhongang Cai, Fangzhou Hong, Huirong Li, Lei Yang, and Ziwei Liu.
\newblock Remodiffuse: Retrieval-augmented motion diffusion model.
\newblock In {\em Proceedings of the IEEE/CVF International Conference on Computer Vision}, pages 364--373, 2023.

\bibitem{zhang2024large}
Mingyuan Zhang, Daisheng Jin, Chenyang Gu, Fangzhou Hong, Zhongang Cai, Jingfang Huang, Chongzhi Zhang, Xinying Guo, Lei Yang, Ying He, et~al.
\newblock Large motion model for unified multi-modal motion generation.
\newblock {\em arXiv preprint arXiv:2404.01284}, 2024.

\bibitem{zhang2024finemogen}
Mingyuan Zhang, Huirong Li, Zhongang Cai, Jiawei Ren, Lei Yang, and Ziwei Liu.
\newblock Finemogen: Fine-grained spatio-temporal motion generation and editing.
\newblock {\em Advances in Neural Information Processing Systems}, 36, 2024.

\bibitem{zhao2023diffugesture}
Weiyu Zhao, Liangxiao Hu, and Shengping Zhang.
\newblock Diffugesture: Generating human gesture from two-person dialogue with diffusion models.
\newblock In {\em Companion Publication of the 25th International Conference on Multimodal Interaction}, pages 179--185, 2023.

\bibitem{zhu2024development}
Yufan Zhu, Silu Chen, Chi Zhang, Zhongyu Piao, and Guilin Yang.
\newblock Development of adaptive safety constraint by predicting trajectories of closest points between human and co-robot.
\newblock {\em Journal of Intelligent Manufacturing}, 35(3):1197--1206, 2024.

\end{thebibliography}

\end{document}